\documentclass{article}

 \usepackage[preprint]{neurips_2026}

\usepackage[utf8]{inputenc} 
\usepackage[T1]{fontenc}    
\usepackage{hyperref}       
\usepackage{url}            
\usepackage{booktabs}       
\usepackage{amsfonts}       
\usepackage{nicefrac}       
\usepackage{microtype}      
\usepackage{xcolor}         
\usepackage{wrapfig}

\usepackage{graphicx}
\usepackage{dblfloatfix}
\usepackage{algorithm}
\usepackage{algpseudocode}
\usepackage{subcaption}
\usepackage{float}
\usepackage{pgfplots}
\pgfplotsset{compat=1.18}
\usepackage{orcidlink}
\usepackage[accsupp]{axessibility}

\title{Aero-World: Action-Conditioned Aerial Video Generation from Inertial Controls
}
\author{%
  Abdul Mohaimen Al Radi \and Kunyang Li \and Yuzhang Shang \\ \and
  Mubarak Shah \and Yu Tian \\
  Institute of Artificial Intelligence, University of Central Florida \\
  \texttt{\{ab575577, kunyang.li, yuzhang.shang, yu.tian2\}@ucf.edu, shah@crcv.ucf.edu}
}

\begin{document}






\maketitle
\begin{abstract}
Foundation video models produce visually impressive results, but their use in embodied AI remains limited because they are primarily trained on natural language rather than low-level control signals. This limitation is especially pronounced for aerial flight, where motion occurs in unconstrained 6-DoF space and small errors in ego-motion can produce large trajectory drift. Generating aerial videos that follow fine-grained inertial actions can support scalable training and evaluation of aerial agents by providing a controllable proxy for real-world or expensive simulation data. To address this problem, we propose \textbf{Aero-World}, a method for converting a pretrained image-to-video diffusion model into a controllable aerial video generator. Aero-World injects sequences of translational acceleration and angular velocity into a pretrained latent diffusion transformer through an action-token stream. A frozen latent-space Physics Probe, trained independently on real video--IMU pairs, provides differentiable inertial-consistency supervision during LoRA finetuning while avoiding computationally expensive video decoding. We further propose \textbf{AeroBench}, a benchmark for evaluating whether generated drone videos adhere to low-level action signals. AeroBench uses Action Alignment Score (AAS) to measure agreement with commanded inertial actions and Physical Consistency Rate (PCR) to measure temporal motion stability. On AeroBench, Aero-World improves mean AAS from 57.7 to 63.6 over action-only finetuning and gives a stronger quality-control trade-off than AirScape, with lower FVD (596.5 vs. 1058.6), higher SSIM (0.595 vs. 0.505), and higher Flow-IMU correlation (0.44 vs. 0.20). These results suggest that frozen Physics Probe supervision is a practical mechanism for adapting pretrained video generators toward more action-aligned aerial motion.
\end{abstract}

\section{Introduction}
\label{sec:intro}

Recent large-scale video diffusion models exhibit impressive perceptual realism~\cite{cog, ltx, hunyuan, opensora, wan}. These models hold the potential to become simulators for embodied agents, which can be utilized for learning policies. However, in order to be a reliable proxy to real-world data or expensive simulations, these models need to possess action-faithfulness and physical consistency, where they are currently lacking~\cite{zhou2026drivinggen, zheng2025vbench}. This gap is more acute in aerial videos because an aerial object like drone operates in unconstrained 6 degree-of-freedom (6-DoF) space, and the translational and rotational dynamics of drones interact non-linearly~\cite{uavcontrol}. Small errors in roll or pitch induce large spatial drift, and aggressive maneuvers amplify temporal inconsistency. A video model which can produce action-faithful aerial videos thus provides a critical benefit: the ability to safely and scalably train robust control policies entirely in software, reducing cost of operating in extreme environments.

\begin{figure}
  \centering
  \includegraphics[width=1\linewidth]{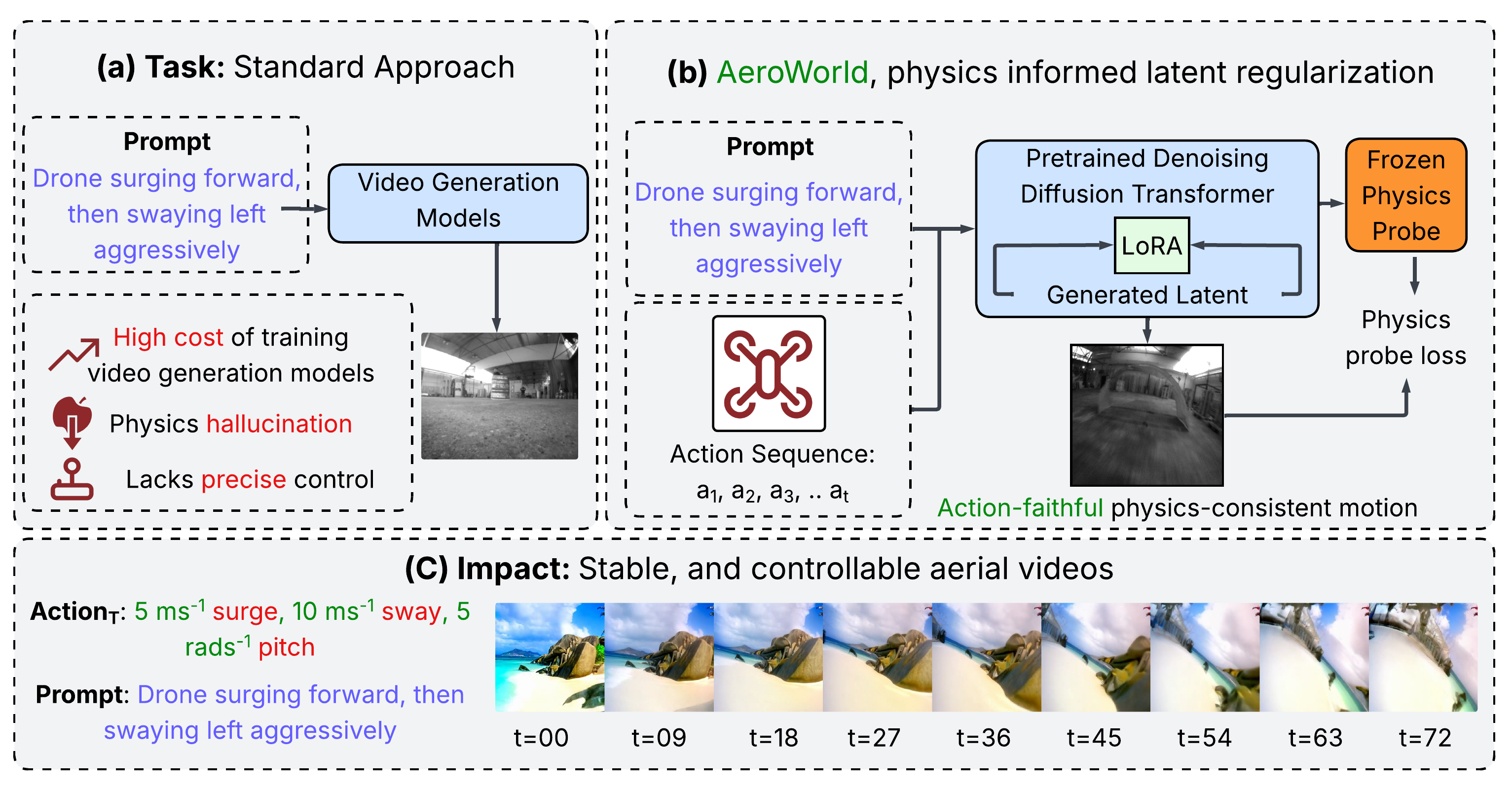}
  \caption{(a) Standard Approach: Language-conditioned video models generate plausible motion without explicit action grounding.
  (b) Aero-World: injects latent-space physics supervision and 6-DoF IMU conditioning into a pretrained diffusion backbone.
  (c) The result is stable, action-faithful aerial motion, even in out-of-distribution scenes.}
  \label{fig:tease}
\end{figure}

In order to solve this, existing literature has introduced simple actions like forward, backwards, left, and right \cite{genie, gaia1, cosmos, dreamerv3}. But this framing has one issue, where there is no way to increase the actuation forces, meaning there is no way to rapidly increase or decrease speed in the ego-motion videos, which is prevalent in most drone flights. Another direction is to add camera parameters, optical flow, pose/depth guidance~\cite{ji2025posetraj, yu2025context, xu2025virtually} to steer the video generation. However, these controls often (i) rely on intermediate annotations or sensors not available to an embodied agent during training~\cite{hao2026egosim}, (ii) do not match the agent's native control interface~\cite{xu2026worldmark}, and (iii) typically lack a training-time objective that explicitly enforces \emph{action-motion consistency} of ego-motion \cite{heng2025imagine2act}.

Recent literature has also confirmed that scaling up data and the model does not correlate well with action adherence \cite{kang2024far, motamed2025generative}. Obtaining annotations for aerial videos is also very difficult and costly. This is why most pretraining corpora consists primarily of captioned videos rather than paired control signals such as linear acceleration and angular velocity. The low-level 6-DoF inertial signals required for control, namely linear acceleration and angular velocity, are rarely observed as supervision in natural video corpora \cite{miech2019howto100m, bain2021frozen, wang2023internvid}.
Thus, scaling primarily improves visual realism, not low-level controllability, motivating explicit action-consistency supervision during adaptation.

To summarize, we have identified generating aerial videos adhering to control signals is an important challenge for embodied AI.

We aim to solve this problem leveraging the existing visual knowledge of video generation models. Rather than retraining a video generation model from scratch, we propose a physics probe-regularized conversion mechanism that reshapes the latent dynamics of a pretrained video diffusion model through inertial-consistency supervision. Because inertial trajectories describe platform-level motion rather than drone-specific motor commands, they provide a control interface that is less tied to a particular actuator model. This also addresses the lack of fine-grained control like rapidly increasing or reducing speed. Working in latent space also means there is no need for heavy decoding during finetuning.

Our key idea is to introduce a frozen physics probe that estimates 6-DoF inertial signals (linear acceleration, and angular velocity) from generated video latents, providing a data-driven surrogate for enforcing motion consistency during finetuning.
During finetuning, the generator is encouraged to produce motion that matches the commanded inertial sequence by penalizing discrepancies between probe-inferred motion and the input actions. 
Importantly, the probe is trained independently on real video–IMU pairs and remains frozen during generator adaptation, providing a stable supervision signal without modifying the pretrained DiT backbone. This mechanism can be attached to any existing diffusion video models with minimal architectural changes, enabling them to generate IMU-conditioned aerial videos while preserving the visual priors learned from large-scale video pretraining.

\begin{figure}[!t]
  \centering
  \includegraphics[width=1.0\linewidth]{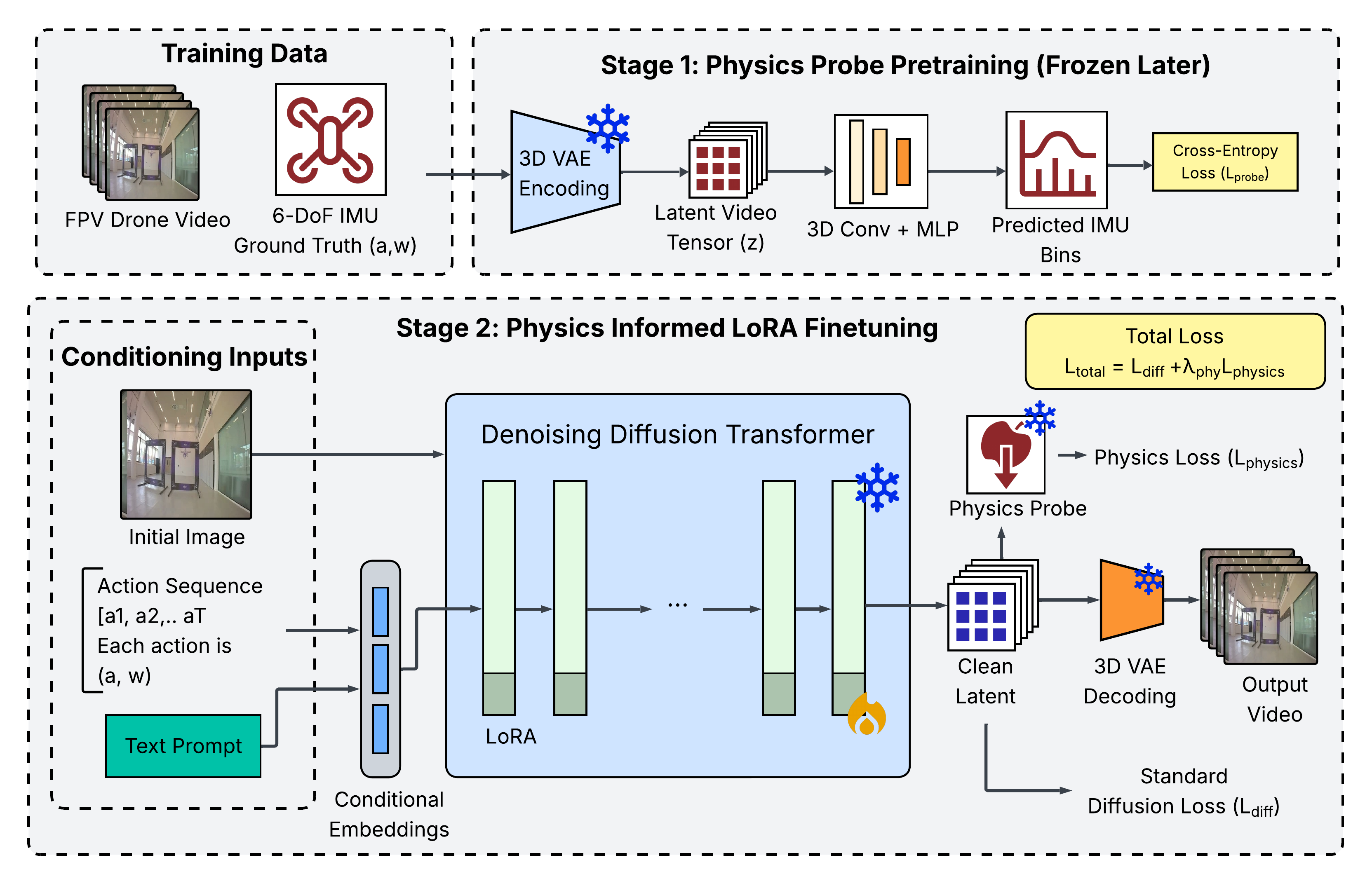}
  \caption{The proposed architecture. A pretrained diffusion backbone generates video latents conditioned on IMU actions.
  A frozen physics probe evaluates latent motion consistency and provides supervision during LoRA finetuning.}
  \label{fig:pipeline}
\end{figure}

Our approach, illustrated in Fig.~\ref{fig:pipeline}, instantiates this conversion on a pretrained diffusion video backbone.
We train on synchronized frames with IMU sequences from drone flight datasets \cite{tii, uhz1}, where high angular velocities and rapid vertical transitions stress temporal coherence.
By injecting action embeddings and applying latent-space inertial-consistency regularization, the model learns to respect 6-DoF motion constraints while preserving perceptual realism inherited from large-scale pretraining.

We evaluate this framework on our proposed \textbf{AeroBench}, a new benchmark designed for action-conditioned aerial generation.
To quantify controllability, we introduce the \textbf{Action Alignment Score (AAS)}, which measures bin-level agreement between commanded IMU actions and probe-inferred motion, and the \textbf{Physical Consistency Rate (PCR)}, which captures frame-to-frame oscillatory instability.

Our contributions are:
\begin{itemize}
    \item \textbf{Inertia-Guided Adaptation:} A method for adapting a pretrained diffusion video backbone for IMU-conditioned aerial video generation using latent-space inertial supervision.

    \item \textbf{Physics Probe Supervision:} A frozen differentiable probe that encourages action–motion consistency during lightweight LoRA finetuning.

    \item \textbf{AeroBench and Metrics:} A benchmark with AAS and PCR for evaluating 6-DoF action alignment and temporal stability in action-conditioned aerial video generation.
\end{itemize}

\section{Related Work}
\noindent\textbf{Foundation Video Models.}
Large-scale video generation has rapidly advanced through diffusion and transformer-based architectures, producing long, high-fidelity samples from text and image prompts \cite{cog, ltx, hunyuan, wan}. Recent systems emphasize improved temporal consistency, stronger video autoencoders, and scalable denoising transformers, enabling high-quality image-to-video generation at large model sizes. However, these models are primarily designed for visual realism and prompt adherence ; they typically provide only coarse control signals like text, style \cite{cog}, or sparse conditioning and do not natively support precise low-level action inputs required for robotics control. Models such as LTX-Video \cite{ltx} exemplify this trend: strong long-horizon generation with limited fine-grained controllability \cite{opensora, sd, latte, phenaki, makeavideo}. These models nevertheless provide strong visual priors, which motivates our approach: rather than training a generator from scratch, we adapt a pretrained video model and guide it toward more action-consistent motion. 

\noindent\textbf{Action-Conditioned and Controllable Video Generation.}
A growing body of work augments video generation with structured controls beyond text, including camera motion tokens \cite{xu2025virtually}, depth/pose guidance \cite{ji2025posetraj}, keypoint trajectories \cite{dragnuwa}, optical-flow conditioning, and other intermediate representations \cite{yu2025context, uavcontrol, zhao2025airscape}. These approaches can provide meaningful steering, but often require additional sensors or annotations, and may not align with the control interfaces used in robotics like IMU-derived 6-DoF velocities. In contrast, Aero-World directly conditions on 6-DoF inertial actions and optimizes the generator to reflect those actions in its visual motion, enabling a control interface that is compatible with embodied agents.

\noindent\textbf{Vision-Language-Action Models and Embodied Foundation Models.}
Vision-language-action (VLA) models and embodied foundation models learn to map sensory observations and language to action sequences, often trained from large-scale demonstrations \cite{openvla, palme, gato, rt2}. While VLA models aim to produce actions, they typically assume an external environment (real or simulated) for interaction. Aero-World is complementary: it focuses on learning an action-conditioned visual generator that produces aerial video sequences aligned with 6-DoF control inputs. Combining VLA policies with controllable generative world models is a promising direction for scalable embodied agent training \cite{dronevla, racevla, dreamerv3}.

\section{Methodology}

\noindent\textbf{Overview.}
We study the problem of converting a pretrained image-to-video latent diffusion model into an aerial video generator controlled by dense inertial commands. Given an initial frame $I_1$ and a 30Hz action sequence
$
a_{1:T}=\{a_1,\ldots,a_T\},\; a_t\in\mathbb{R}^6,
$
where $a_t=(\alpha_x,\alpha_y,\alpha_z,\omega_x,\omega_y,\omega_z)$ contains translational acceleration and angular velocity, the goal is to generate a rollout $\hat{V}_{1:T}$ whose ego-motion follows the commanded inertial trajectory.

A key difficulty is that standard diffusion finetuning can reconstruct training videos without necessarily using the action signal; the denoising loss alone does not explicitly penalize action-motion mismatch. Aero-World therefore combines two mechanisms. First, an action-token stream embeds IMU commands and injects them into the diffusion transformer, giving the generator a low-level control interface. Second, a frozen latent-space Physics Probe provides a differentiable motion-consistency signal during finetuning. The probe is trained independently on real synchronized video--IMU pairs, then kept fixed so that generator adaptation is guided by an external inertial evaluator rather than a co-adapted loss network. At inference time, the probe is discarded and the model generates videos using only the initial frame and IMU action sequence.

\subsection{Physics Probe}
\label{sec:physics_probe}

The Physics Probe $\Phi$ is a frozen latent-space evaluator that maps video motion to inertial states. Its purpose is not to replace the generator, but to provide a differentiable training signal indicating whether the generated latent trajectory contains motion consistent with the commanded IMU sequence. We train the probe before generator finetuning using only real synchronized video--IMU pairs.

Let $z=\mathcal{E}(V)\in\mathbb{R}^{B\times C\times T_\ell\times H_\ell\times W_\ell}$ denote the latent tensor produced by the frozen video autoencoder. For each IMU axis $j$, a lightweight probe $\Phi_j$ predicts per-timestep logits
$
\Phi_j(z)_t\in\mathbb{R}^{K}
$
over $K$ discretized inertial bins. We use axis-specific prediction heads because the six inertial dimensions have different dynamic ranges and class imbalance patterns. The probe architecture consists of lightweight 3D convolutional layers over latent space, global spatial pooling, and an MLP classifier. The 3D convolutions preserve local spatio-temporal inductive bias while avoiding the cost of decoding generated videos back to RGB during finetuning.

Rather than regressing continuous IMU values, we discretize each axis into $K$ bins over an axis-specific physical range $[m_j,M_j]$. The bin label is
\begin{equation}
\label{eq:quant}
\mathcal{Q}_j(x)=
\min\left(
K-1,
\left\lfloor
\frac{\mathrm{clip}(x,m_j,M_j)-m_j}{M_j-m_j}\cdot K
\right\rfloor
\right).
\end{equation}

Binned supervision is used for two reasons. First, aggressive drone flight produces heavy-tailed IMU distributions, making direct regression sensitive to rare high-magnitude maneuvers. Second, for action-conditioned generation, the model mainly needs to recover the correct motion regime, such as accelerating, turning, or yawing, rather than sensor-grade IMU reconstruction. Given ground-truth bin labels $y^{(j)}_t=\mathcal{Q}_j(s^{(j)}_t)$, the probe is trained with
\begin{equation}
\label{eq:probe}
\mathcal{L}_{probe}
=
\frac{1}{6T}
\sum_{j=1}^{6}\sum_{t=1}^{T}
\mathrm{CE}\!\left(\Phi_j(z)_t,y^{(j)}_t\right)
\end{equation}
After training, $\Phi$ is frozen. During generator adaptation, gradients are allowed to pass through $\Phi$ to the generated latent prediction, but the probe parameters themselves are never updated.

\subsection{Probe-Regularized LoRA Finetuning}
\label{sec:lora_physics}

We adapt the pretrained diffusion backbone using Low-Rank Adaptation (LoRA), while keeping the original backbone weights and the Physics Probe frozen. The trainable components are the LoRA adapters and the action embedding layers. Each action vector $a_t$ is embedded by a lightweight MLP and injected into transformer blocks through cross-attention, providing a separate action-token stream in addition to visual and timestep conditioning.

We use a causal action alignment: the command $a_{t+1}$ conditions the transition into frame $\hat{I}_{t+1}$. This shift is important because IMU measurements describe changes in motion between consecutive frames rather than the static appearance of the current frame. Without this alignment, the model may learn a temporally offset mapping between actions and visual motion.

Given a real training clip $V$, we encode it into clean latents $z_0=\mathcal{E}(V)$, sample diffusion noise $\epsilon$ and timestep $t$, and optimize the standard denoising loss

\begin{equation}
    \mathcal{L}_{diff}
=
\mathbb{E}_{z_0,\epsilon,t}
\left[
\|\epsilon-\epsilon_\theta(z_t,a_{1:T})\|_2^2
\right]
\end{equation}

with the final loss being \(L_{total} = L_{diff} + \lambda_{phy}L_{probe}\).
The denoising objective teaches the model to generate realistic aerial video latents, but does not by itself enforce that the generated motion uses the provided IMU commands. We therefore recover the denoised latent prediction $\hat{z}_0$ from the model output and evaluate it with the frozen probe. The probe is applied to $\hat{z}_0$ rather than the noisy latent $z_t$ because it was trained on clean VAE latents from real videos.

For each IMU axis $j$, the probe predicts logits $\Phi_j(\hat{z}_0)_t$, which are compared against the commanded action bin $y^{(j)}_t$ is same as equation \ref{eq:probe}.
This objective preserves the visual prior of the pretrained model through $\mathcal{L}_{diff}$, while $\mathcal{L}_{phys}$ biases the LoRA update toward latent trajectories whose inferred inertial motion matches the input command. We use a warm-up stage with $\lambda_{phys}=0$ before enabling probe supervision, which stabilizes training by first teaching the newly introduced action pathway to interact with the pretrained backbone.
\section{Experiments}

We evaluate Aero-World in terms of visual fidelity, action alignment, and temporal stability. 
This section first describes the experimental setup and training configuration, then we introduce our AeroBench benchmark, followed by the evaluation metrics and both quantitative and qualitative results.

\subsection{Experimental Setup}

\noindent\textbf{Dataset.}
We use the UZH FPV drone racing dataset \cite{uhz1, uzh2, uzh3} and the TII drone racing dataset \cite{tii}, both providing synchronized RGB video and IMU measurements. 
We focus on FPV footage to preserve raw ego-motion without camera stabilization artifacts.

Videos are resampled to 30\,Hz while IMU signals (originally 500\,Hz) are temporally aligned to the video frames. 
Each sequence is clipped to 81 frames to match the diffusion context window and to reduce repetitive motion patterns in the TII dataset. 
The resulting dataset contains 7k training clips, 1k validation clips, and 1k test clips.

\noindent\textbf{Physics probe training.}
The Physics Probe is trained using cached VAE latents paired with synchronized IMU sequences. We chose CogVideoX-1.5-5B I2V \cite{cog} as our diffusion transformer backbone. Its 3D VAE encoder is frozen and produces latent tensors with $C=16$ channels for clips of length $T=81$. 
We train one lightweight probe per IMU axis because the six inertial dimensions have different dynamic ranges and class imbalance patterns.

All probes are trained using AdamW with learning rate $10^{-4}$ and batch size 32 for 25 epochs under bfloat16 mixed precision. 
Model checkpoints are selected based on validation accuracy.

\begin{wrapfigure}{r}{0.6\textwidth}
    \centering
    \begin{tikzpicture}
    \begin{axis}[
    width=0.86\linewidth,
    height=6.7cm,
    ybar,
    bar width=7pt,
    ymin=0, ymax=100,
    ylabel={Accuracy (\%)},
    symbolic x coords={
        {$\alpha_x$},{$\alpha_y$},{$\alpha_z$},
        {$\omega_x$},{$\omega_y$},{$\omega_z$}
    },
    xtick=data,
    x tick label style={font=\small},
    y tick label style={font=\small},
    label style={font=\small},
    enlarge x limits=0.16,
    grid=major,
    grid style={dotted},
    clip=false,
    legend style={
        font=\small,
        at={(1.02,1.00)},
        anchor=north west,
        draw=none,
        fill=none,
        row sep=2pt
    },
    legend cell align=left,
    nodes near coords,
    nodes near coords align={vertical},
    every node near coord/.append style={
        font=\scriptsize,
        yshift=0pt
    },
]

        \addplot coordinates {
            ($\alpha_x$,14.29) ($\alpha_y$,14.29) ($\alpha_z$,14.29)
            ($\omega_x$,14.29) ($\omega_y$,14.29) ($\omega_z$,14.29)
        };
        \addlegendentry{Random (1/7)}

        \addplot coordinates {
            ($\alpha_x$,45.10) ($\alpha_y$,66.10) ($\alpha_z$,32.00)
            ($\omega_x$,68.80) ($\omega_y$,80.50) ($\omega_z$,63.00)
        };
        \addlegendentry{Majority bin}

        \addplot coordinates {
            ($\alpha_x$,80.23) ($\alpha_y$,77.07) ($\alpha_z$,67.62)
            ($\omega_x$,89.35) ($\omega_y$,94.32) ($\omega_z$,87.91)
        };
        \addlegendentry{Physics Probe}

    \end{axis}
    \end{tikzpicture}
     \caption{\textbf{Physics Probe accuracy vs.\ baselines.} Per-axis classification accuracy over $K{=}7$ discretized bins. The blue indicates the accuracy of choosing a random bin uniformly, the red indicates the accuracy if always the majority bin is chosen. The Physics Probe substantially outperforms both random and majority-bin baselines across all six axes.}
    \label{fig:probe_baselines}
\end{wrapfigure}
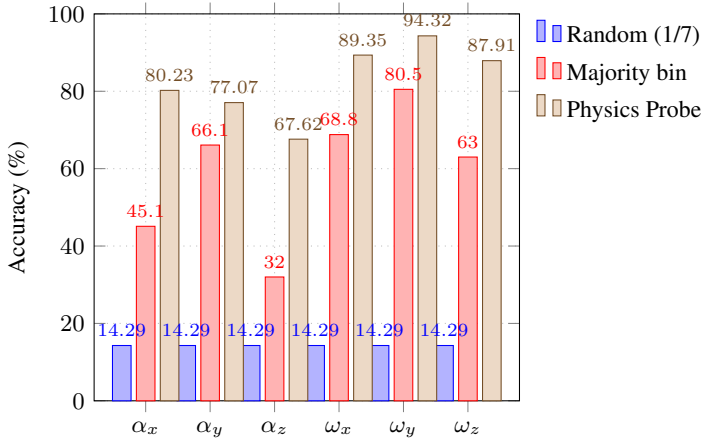
\noindent\textbf{LoRA finetuning.}
We finetune the diffusion backbone using Low-Rank Adaptation (LoRA) \cite{lora}, keeping the pretrained model and the Physics Probe frozen. 
Training proceeds in two stages: a warm-up phase with $\lambda_{phys}=0$ for 10k steps, followed by joint optimization with $\lambda_{phys}=0.2$. We also show sensitivity of $\lambda_{phys}$ and performance of different DiT backbones to show generalization in the supplementary material.
The model is trained for 70k steps using 2 H100 GPUs with batch size 8 and 81-frame clips.

All baselines use the same test split, initial frames, prompts, 81-frame horizon, and metric code; Cosmos is evaluated zero-shot, while AirScape is evaluated through its released motion-intention interface using the closest IMU-to-intention conversion without rejection sampling.

\subsection{Evaluation Metrics}

\noindent\textbf{Action Alignment Score (AAS).}
Action-faithfulness is measured using a probe-based Action Alignment Score (AAS). 
For each IMU axis $j$, a frozen probe $\Phi_j$ predicts a $K$-way discretized bin from VAE latents. 
Continuous IMU values are quantized as equation~\ref{eq:quant}. Actions are causally aligned by left-shifting the IMU sequence to match the transition $t \rightarrow t+1$. 
Given a generated video $\hat{V}$, the alignment score is

\[
\mathrm{AAS}^{(j)}=
\frac{1}{T}\sum_{t=1}^{T}
\mathbf{1}\!\left[
\arg\max \Phi_j(\mathcal{E}(\hat{V}))_t
=
\mathcal{Q}_j(\tilde{s}_t)
\right].
\]

We report per-axis AAS and the mean across axes.

\noindent\textbf{Physical Consistency Rate (PCR).}
While AAS measures bin-level alignment, it does not capture temporal stability. 
We therefore define PCR using first-order derivatives of probe-predicted IMU bins. 
For each axis $j$,
Let
$
\hat{s}^{(j)}_t=\arg\max \Phi_j(\mathcal{E}(\hat{V}))_t
$
denote the probe-predicted IMU bin for axis $j$ at frame $t$. We define
\[
\mathrm{PCR}^{(j)}
=
\frac{1}{T-1}
\sum_{t=2}^{T}
\left|
\hat{s}^{(j)}_t-\hat{s}^{(j)}_{t-1}
\right|.
\]
Lower PCR indicates smoother probe-inferred inertial motion.

PCR is defined as the mean absolute bin jump between consecutive frames; lower values indicate smoother motion.
\subsection{Quantitative Results}

We evaluate Aero-World along two dimensions: visual fidelity and physical controllability. We test both visuals and physical controllability to show the trade-off between them.

\noindent\textbf{Visual fidelity.}
Figure~\ref{fig:visual_fidelity_chart} reports FVD and SSIM. Action finetune(FT) achieves the lowest FVD, but weaker controllability. AirScape obtains competitive AAS/PCR, but its videos are strongly off-distribution on AeroBench (FVD 1058.6, SSIM 0.505), so we do not interpret its control scores in isolation. Aero-World gives a better usable trade-off than AirScape \cite{zhao2025airscape}: lower FVD, higher SSIM, higher AAS, and higher Flow-IMU.

\noindent\textbf{Action alignment.}
As defined before, AAS measures how aligned the observed actions are with the given input action. Figure ~\ref{fig:triple_controllability_analysis}(a) reports per-axis AAS. We observe that action conditioning enables measurable 6-DoF alignment, while physics supervision improves mean AAS from 57.67 to \textbf{63.62}. We observe cosmos \cite{cosmos} having similar temporal stability, but worse visual fidelity as well as action alignment.

\noindent\textbf{Temporal stability.}
Figure~\ref{fig:triple_controllability_analysis}(b) reports PCR. PCR measures how abruptly the probe-inferred inertial bins change between consecutive frames. A sudden change will mean jittery video or unrealistic trajectory. Physics supervision slightly reduces PCR (0.033 $\rightarrow$ 0.029), indicating smoother frame-to-frame dynamics. We see the PCR is marginally better for linear acceleration in x-axis for the variant without physics supervision suggesting that physics supervision does not uniformly improve smoothness across all motion axes.

\begin{figure}[htbp]
    \centering
    \begin{tikzpicture}
    \begin{axis}[
        width=0.95\linewidth,
        height=5cm,
        ybar=2pt,
        bar width=10pt,
        symbolic x coords={Pretrained, Cosmos, AirScape, Action FT, Ours},
        xtick=data,
        enlarge x limits=0.15,
        ymin=0, ymax=1200,
        ylabel={FVD $\downarrow$},
        ylabel style={font=\small, color=blue!70!black},
        yticklabel style={font=\small, color=blue!70!black},
        ymajorgrids=true,
        grid style={dashed, gray!30},
        legend style={at={(0.5,-0.2)}, anchor=north, legend columns=2, draw=none, font=\small},
    ]
        \addplot[fill=blue!30, draw=blue!60] coordinates {
            (Pretrained,698.6) (Cosmos,874.5) (AirScape,1058.6) (Action FT,504.9) (Ours,596.5)
        };
        \addlegendentry{FVD (Quality)}

        \addplot[
            color=red!70!black,
            mark=square*,
            thick,
            axis y line=right, 
        ] coordinates {
            (Pretrained,520.7) (Cosmos,598.9) (AirScape,504.9) (Action FT,600.1) (Ours,595.4)
        };
        \addlegendentry{SSIM $\times 10^3$ (Similarity)}
        
    \end{axis}

    \begin{axis}[
        width=0.95\linewidth,
        height=5cm,
        axis y line=right,
        axis x line=none,
        ymin=450, ymax=650, 
        ylabel={SSIM $\uparrow$},
        ylabel style={font=\small, color=red!70!black},
        yticklabel style={font=\small, color=red!70!black},
        enlarge x limits=0.15,
        symbolic x coords={Pretrained, Cosmos, AirScape, Action FT, Ours},
    ]
    \end{axis}
    \end{tikzpicture}
    \caption{\textbf{Visual Fidelity Trade-off.} While action-only finetuning achieves the lowest FVD, our physics-regularized model (Ours) maintains superior perceptual quality compared to base models and SOTA competitors like AirScape, without sacrificing structural similarity (SSIM).}
    \label{fig:visual_fidelity_chart}
\end{figure}
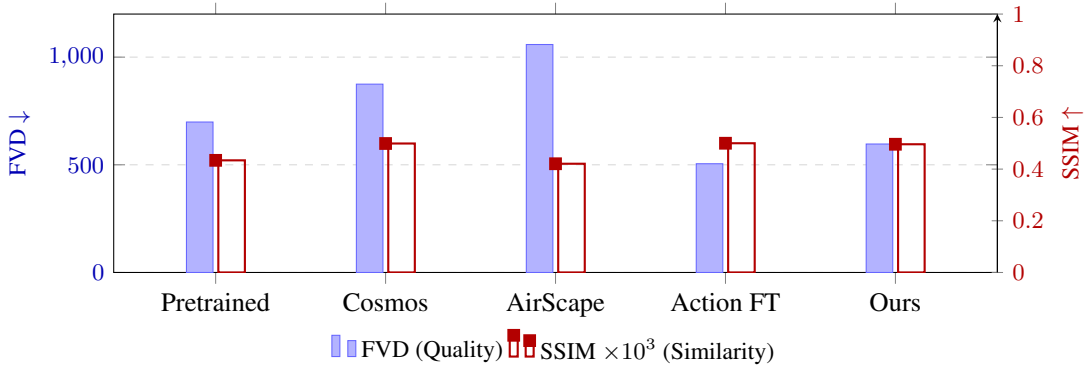
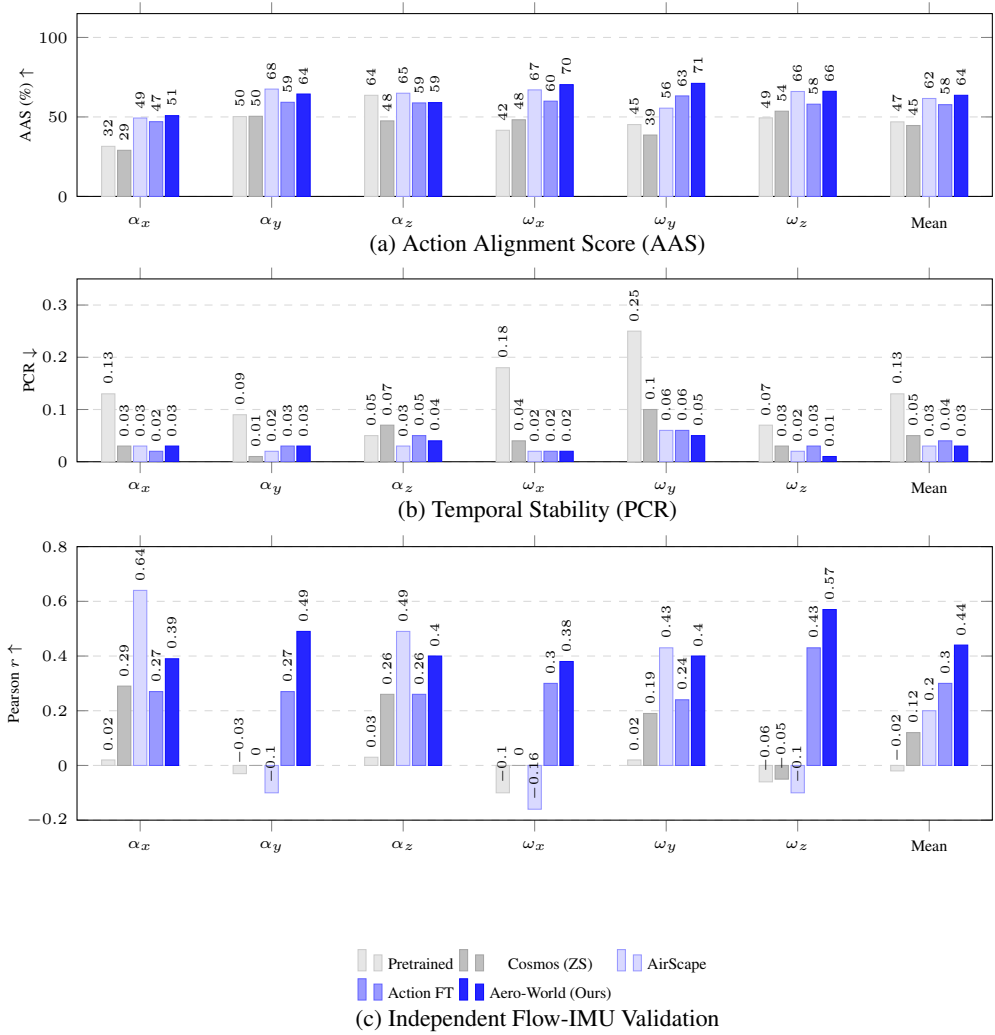
\begin{figure}[t]
    \centering
    \begin{subfigure}{\linewidth}
        \centering
        \begin{tikzpicture}[trim axis left, trim axis right]
        \begin{axis}[
            width=0.98\linewidth, height=4cm,
            ybar=1pt, bar width=5pt,
            ymin=0, ymax=115,
            ylabel={AAS (\%) $\uparrow$},
            ylabel style={font=\tiny, yshift=-5pt},
            symbolic x coords={$\alpha_x$, $\alpha_y$, $\alpha_z$, $\omega_x$, $\omega_y$, $\omega_z$, Mean},
            xtick=data,
            xticklabel style={font=\tiny},
            yticklabel style={font=\tiny},
            enlarge x limits=0.08,
            ymajorgrids=true, grid style={dashed, gray!30},
            nodes near coords,
            every node near coord/.append style={font=\fontsize{4}{5}\selectfont, rotate=90, anchor=west, /pgf/number format/precision=0}
        ]
            \addplot[fill=gray!20, draw=gray!40] coordinates {($\alpha_x$,31.5) ($\alpha_y$,50.2) ($\alpha_z$,63.6) ($\omega_x$,41.6) ($\omega_y$,45.2) ($\omega_z$,49.4) (Mean,46.9)};
            \addplot[fill=gray!50, draw=gray!70] coordinates {($\alpha_x$,29.0) ($\alpha_y$,50.4) ($\alpha_z$,47.5) ($\omega_x$,48.2) ($\omega_y$,38.6) ($\omega_z$,53.6) (Mean,44.6)};
            \addplot[fill=blue!15, draw=blue!40] coordinates {($\alpha_x$,49.3) ($\alpha_y$,67.5) ($\alpha_z$,64.9) ($\omega_x$,67.0) ($\omega_y$,55.5) ($\omega_z$,66.0) (Mean,61.6)};
            \addplot[fill=blue!40, draw=blue!60] coordinates {($\alpha_x$,47.0) ($\alpha_y$,59.2) ($\alpha_z$,58.8) ($\omega_x$,59.9) ($\omega_y$,63.2) ($\omega_z$,58.0) (Mean,57.7)};
            \addplot[fill=blue!85, draw=blue!95] coordinates {($\alpha_x$,50.8) ($\alpha_y$,64.4) ($\alpha_z$,59.0) ($\omega_x$,70.3) ($\omega_y$,71.1) ($\omega_z$,66.1) (Mean,63.6)};
        \end{axis}
        \end{tikzpicture}
        \vspace{-8pt}
        \caption{Action Alignment Score (AAS)}
    \end{subfigure}

    \vspace{2pt}

    \begin{subfigure}{\linewidth}
        \centering
        \begin{tikzpicture}[trim axis left, trim axis right]
        \begin{axis}[
            width=0.98\linewidth, height=4cm,
            ybar=1pt, bar width=5pt,
            ymin=0, ymax=0.35,
            ylabel={PCR $\downarrow$},
            ylabel style={font=\tiny, yshift=-5pt},
            symbolic x coords={$\alpha_x$, $\alpha_y$, $\alpha_z$, $\omega_x$, $\omega_y$, $\omega_z$, Mean},
            xtick=data,
            xticklabel style={font=\tiny},
            yticklabel style={font=\tiny},
            enlarge x limits=0.08,
            ymajorgrids=true, grid style={dashed, gray!30},
            nodes near coords,
            every node near coord/.append style={font=\fontsize{4}{5}\selectfont, rotate=90, anchor=west, /pgf/number format/fixed, /pgf/number format/precision=2}
        ]
            \addplot[fill=gray!20, draw=gray!40] coordinates {($\alpha_x$,0.13) ($\alpha_y$,0.09) ($\alpha_z$,0.05) ($\omega_x$,0.18) ($\omega_y$,0.25) ($\omega_z$,0.07) (Mean,0.13)};
            \addplot[fill=gray!50, draw=gray!70] coordinates {($\alpha_x$,0.03) ($\alpha_y$,0.01) ($\alpha_z$,0.07) ($\omega_x$,0.04) ($\omega_y$,0.10) ($\omega_z$,0.03) (Mean,0.05)};
            \addplot[fill=blue!15, draw=blue!40] coordinates {($\alpha_x$,0.03) ($\alpha_y$,0.02) ($\alpha_z$,0.03) ($\omega_x$,0.02) ($\omega_y$,0.06) ($\omega_z$,0.02) (Mean,0.03)};
            \addplot[fill=blue!40, draw=blue!60] coordinates {($\alpha_x$,0.02) ($\alpha_y$,0.03) ($\alpha_z$,0.05) ($\omega_x$,0.02) ($\omega_y$,0.06) ($\omega_z$,0.03) (Mean,0.04)};
            \addplot[fill=blue!85, draw=blue!95] coordinates {($\alpha_x$,0.03) ($\alpha_y$,0.03) ($\alpha_z$,0.04) ($\omega_x$,0.02) ($\omega_y$,0.05) ($\omega_z$,0.01) (Mean,0.03)};
        \end{axis}
        \end{tikzpicture}
        \vspace{-8pt}
        \caption{Temporal Stability (PCR)}
    \end{subfigure}

    \vspace{2pt}

    \begin{subfigure}{\linewidth}
        \centering
        \begin{tikzpicture}[trim axis left, trim axis right]
        \begin{axis}[
            width=0.98\linewidth, height=5.2cm,
            ybar=1pt, bar width=5pt,
            ymin=-0.2, ymax=0.8, 
            ylabel={Pearson $r$ $\uparrow$},
            ylabel style={font=\tiny, yshift=-5pt},
            symbolic x coords={$\alpha_x$, $\alpha_y$, $\alpha_z$, $\omega_x$, $\omega_y$, $\omega_z$, Mean},
            xtick=data,
            xticklabel style={font=\tiny},
            yticklabel style={font=\tiny},
            enlarge x limits=0.08,
            ymajorgrids=true, grid style={dashed, gray!30},
            legend style={at={(0.5,-0.45)}, anchor=north, legend columns=3, draw=none, font=\tiny},
            nodes near coords,
            every node near coord/.append style={font=\fontsize{4}{5}\selectfont, rotate=90, anchor=west, /pgf/number format/fixed, /pgf/number format/precision=2}
        ]
            \addplot[fill=gray!20, draw=gray!40] coordinates {($\alpha_x$,0.02) ($\alpha_y$,-0.03) ($\alpha_z$,0.03) ($\omega_x$,-0.10) ($\omega_y$,0.02) ($\omega_z$,-0.06) (Mean,-0.02)};
            \addlegendentry{Pretrained}
            \addplot[fill=gray!50, draw=gray!70] coordinates {($\alpha_x$,0.29) ($\alpha_y$,-0.00) ($\alpha_z$,0.26) ($\omega_x$,0.00) ($\omega_y$,0.19) ($\omega_z$,-0.05) (Mean,0.12)};
            \addlegendentry{Cosmos (ZS)}
            \addplot[fill=blue!15, draw=blue!40] coordinates {($\alpha_x$,0.64) ($\alpha_y$,-0.10) ($\alpha_z$,0.49) ($\omega_x$,-0.16) ($\omega_y$,0.43) ($\omega_z$,-0.10) (Mean,0.20)};
            \addlegendentry{AirScape}
            \addplot[fill=blue!40, draw=blue!60] coordinates {($\alpha_x$,0.27) ($\alpha_y$,0.27) ($\alpha_z$,0.26) ($\omega_x$,0.30) ($\omega_y$,0.24) ($\omega_z$,0.43) (Mean,0.30)};
            \addlegendentry{Action FT}
            \addplot[fill=blue!85, draw=blue!95] coordinates {($\alpha_x$,0.39) ($\alpha_y$,0.49) ($\alpha_z$,0.40) ($\omega_x$,0.38) ($\omega_y$,0.40) ($\omega_z$,0.57) (Mean,0.44)};
            \addlegendentry{Aero-World (Ours)}
        \end{axis}
        \end{tikzpicture}
        \vspace{-8pt}
        \caption{Independent Flow-IMU Validation}
    \end{subfigure}

    \vspace{22pt} 
    \caption{\textbf{Quantitative Benchmarking.} Aero-World (Ours) improves mean action alignment and independent RGB-space Flow-IMU correlation, while maintaining low temporal instability compared with action-only finetuning.}
    \label{fig:triple_controllability_analysis}
\end{figure}
\subsection{Auxiliary Independent Flow-IMU Validation}
\label{sec:flow_imu}
To reduce probe-circularity, we introduce Flow-IMU, an independent RGB-space evaluator that maps decoded optical-flow features to 6-DoF IMU signals using a ridge regressor trained only on real video-IMU pairs. It shares no architecture with the latent probe and is never used during generator finetuning. Since monocular RGB-to-IMU estimation is ill-posed, Flow-IMU is a relative proxy rather than sensor-grade reconstruction.

Figure~\ref{fig:triple_controllability_analysis}(c) shows that Aero-World achieves the highest mean Flow-IMU correlation (0.44), outperforming Action FT (0.30) and AirScape (0.20). AirScape has negative correlations on several axes, whereas Aero-World is positive on all six axes, suggesting that AAS gains also appear in decoded RGB motion.


\subsection{Qualitative Results}
For qualitative results, we show both results from our validation set and from unseen images. 

\begin{figure}[t]
    \centering

    \begin{subfigure}{\linewidth}
        \centering
        \includegraphics[width=0.9\linewidth]{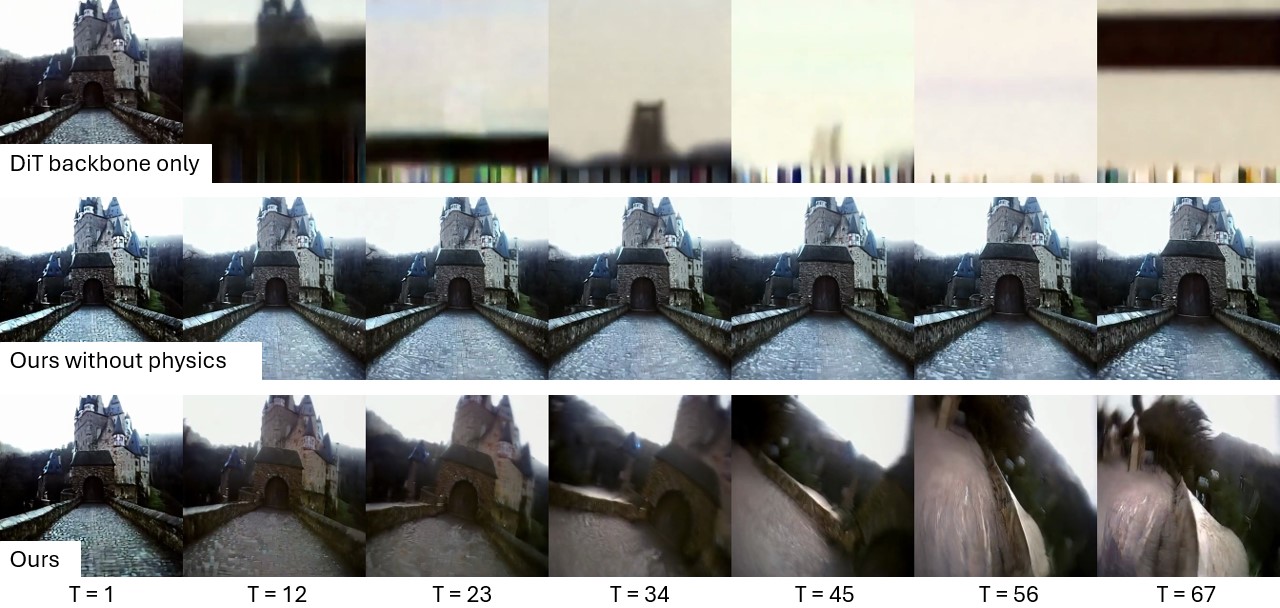}
        \caption{\textbf{Out-of-distribution action-controlled flight.}
        Prompt: \emph{``A drone surging forward fast then swaying left aggressively.''}
        Rows correspond to model variants: \textbf{Top:} caption-conditioned backbone,
        \textbf{Middle:} action-only finetuning,
        \textbf{Bottom:} Aero-World.}
        \label{fig:ood2}
    \end{subfigure}

    \vspace{-0.001em}

    \begin{subfigure}{\linewidth}
        \centering
        \includegraphics[width=.9\linewidth]{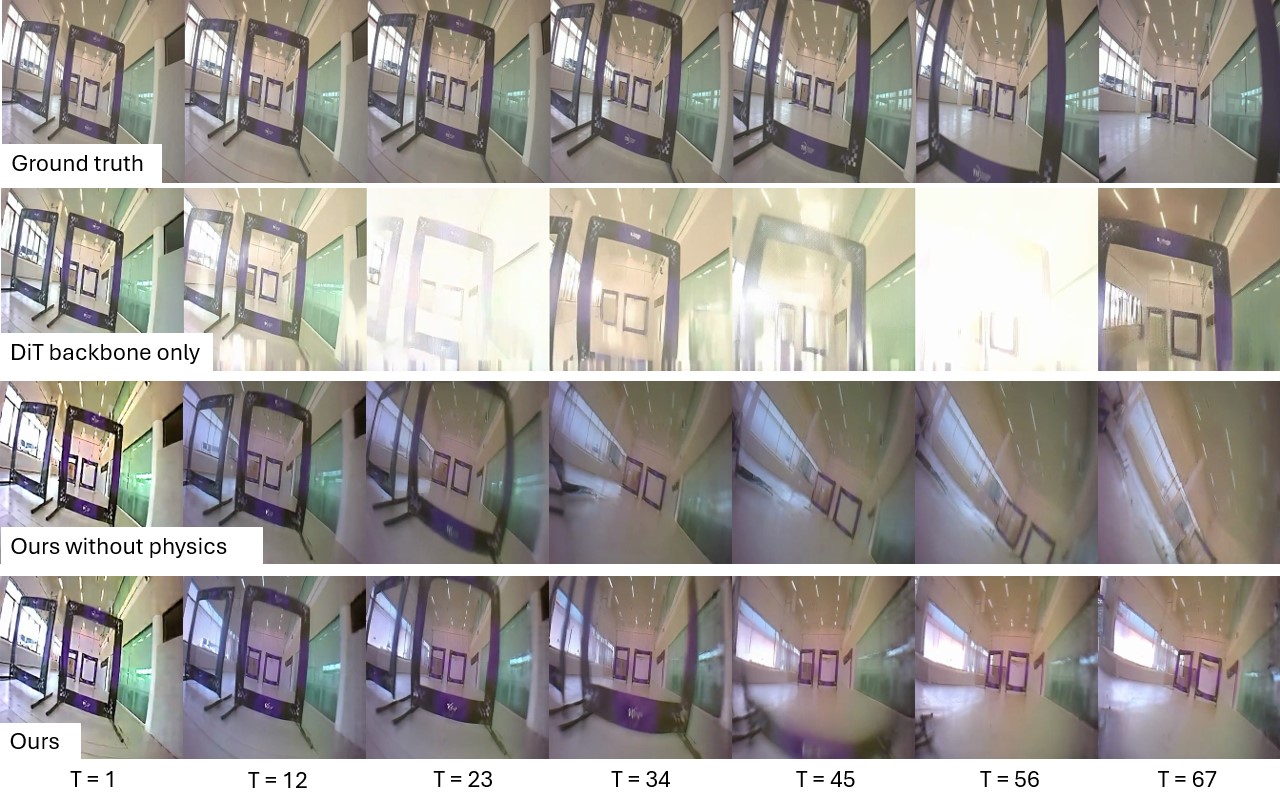}
        \caption{\textbf{In-distribution action-controlled flight.}
        Seven uniformly spaced frames from 81-frame validation rollouts. The top row shows
        the ground-truth video, while the generated sequence follows the commanded inertial trajectory.}
        \label{fig:id1}
    \end{subfigure}

    \vspace{-0.5em}

    \caption{\textbf{Qualitative action-controlled flight results.}
    We show seven uniformly spaced frames from 81-frame rollouts. Aero-World produces stable,
    action-faithful motion in both unseen environments and validation-set maneuvers. Full prompts
    and videos are provided in the supplementary material.}
    \label{fig:qualitative_results}
\end{figure}

\noindent\textbf{Out-of-distribution scenes.}
Figure ~\ref{fig:ood2} demonstrate generation in unseen environments. 
Despite the novel scene distribution, Aero-World preserves stable ego-motion and action-faithful dynamics.

\noindent\textbf{In-distribution maneuvers.}
Figure ~\ref{fig:id1} show validation roll-outs under complex action prompts. 
The generated trajectories follow the commanded inertial signals while maintaining stable orientation.

Full-resolution videos and additional prompts are provided in the \textbf{supplementary material}, as well as details of converting the IMU values into meaningful captions.

\section{Conclusion}
We presented \textbf{Aero-World}, a lightweight framework for adapting pretrained video diffusion models to generate aerial videos conditioned on dense 6-DoF IMU trajectories. By combining explicit action conditioning with a frozen latent Physics Probe during LoRA finetuning, Aero-World improves inertial alignment over matched action-conditioned finetuning, with gains also reflected by an independent RGB-space Flow-IMU evaluator. These results suggest that frozen probe supervision is a practical way to make pretrained video generators more responsive to low-level aerial motion controls, while highlighting a remaining fidelity-control trade-off for future work.
\clearpage

%
%
\bibliographystyle{unsrtnat}
\bibliography{main}


\end{document}